\documentclass[conference]{IEEEtran}
\usepackage{times}
\usepackage{amsmath}
\usepackage{amssymb}
\usepackage{amsthm}
\usepackage{epsfig}
\usepackage{epstopdf}

\usepackage[numbers]{natbib}
\usepackage{multicol}
\usepackage{algorithm}
\usepackage[noend]{algorithmic}

\DeclareMathOperator*{\argmin}{argmin}

\begin{document}

\title{Kinodynamic RRT*: Optimal Motion Planning for Systems with Linear Differential Constraints}

\author{Dustin J. Webb, Jur van den Berg}



%

\maketitle

\begin{abstract}
We present Kinodynamic RRT*, an incremental sampling-based approach for asymptotically optimal motion planning for robots with linear differential constraints. Our approach extends RRT*, which was introduced for holonomic robots \cite{karaman11}, by using a fixed-final-state-free-final-time controller that exactly and optimally connects any pair of states, where the cost function is expressed as a trade-off between the duration of a trajectory and the expended control effort. Our approach generalizes earlier work on extending RRT* to kinodynamic systems, as it guarantees asymptotic optimality for any system with controllable linear dynamics, in state spaces of any dimension. Our approach can be applied to non-linear dynamics as well by using their first-order Taylor approximations. In addition, we show that for the rich subclass of systems with a nilpotent dynamics matrix, closed-form solutions for optimal trajectories can be derived, which keeps the computational overhead of our algorithm compared to traditional RRT* at a minimum.  We demonstrate the potential of our approach by computing asymptotically optimal trajectories in three challenging motion planning scenarios: (i) a planar robot with a 4-D state space and double integrator dynamics, (ii) an aerial vehicle with a 10-D state space and linearized quadrotor dynamics, and (iii) a car-like robot with a 5-D state space and non-linear dynamics.
\end{abstract}

\IEEEpeerreviewmaketitle

\section{Introduction}
Much progress has been made in the area of motion planning in robotics over the past decades, where the basic problem is defined as finding a trajectory for a robot between a start state and a goal state without collisions with obstacles in the environment.  The introduction of incremental sampling-based planners, such as probabilistic roadmaps (PRM) \cite{kavraki96} and rapidly-exploring random trees (RRT) \cite{lavalle01} enabled solving motion planning problems in high-dimensional state spaces in reasonable computation time, even though the problem is known to be PSPACE-hard \cite{latombe91}. PRM and RRT are  asymptotically complete, which means that a solution will be found (if one exists) with a probability approaching 1 if one lets the algorithm run long enough. More recently, an extension of RRT called RRT* \cite{karaman11} was developed that achieves \emph{asymptotic optimality}, which means that an \emph{optimal} solution will be found with a probability approaching 1. 
 
While RRT* has successfully been applied in practice \cite{perez11}, a key limitation of RRT* is that it is applicable only to systems with simple dynamics, as it relies on the ability to connect any pair of states with an optimal trajectory (e.g. holonomic robots, for which straight lines through the state space represent feasible motions). For \emph{kinodynamic} systems, however, straight-line connections between pairs of states are typically not valid trajectories due to the system's \emph{differential constraints}. Finding a feasible trajectory between two states for differentially constrained systems is known as the \emph{two-point boundary value problem} \cite{lavalle06}, and is non-trivial to solve in general. Numerical approaches, such as the shooting method \cite{burden01, jeon11}, are computationally intensive and their solutions may not satisfy any notion of optimality. Prior works on extending RRT* for kinodynamic systems have therefore focused on simple specific instances of kinodynamic systems \cite{karaman10b}, or have serious limitations as they do not in fact succeed to compute an optimal trajectory between any pair of states \cite{jeon11, perez12} (see our discussion in Section \ref{sec:relatedwork}). 

\begin{figure}
\centering
\includegraphics[width=200pt]{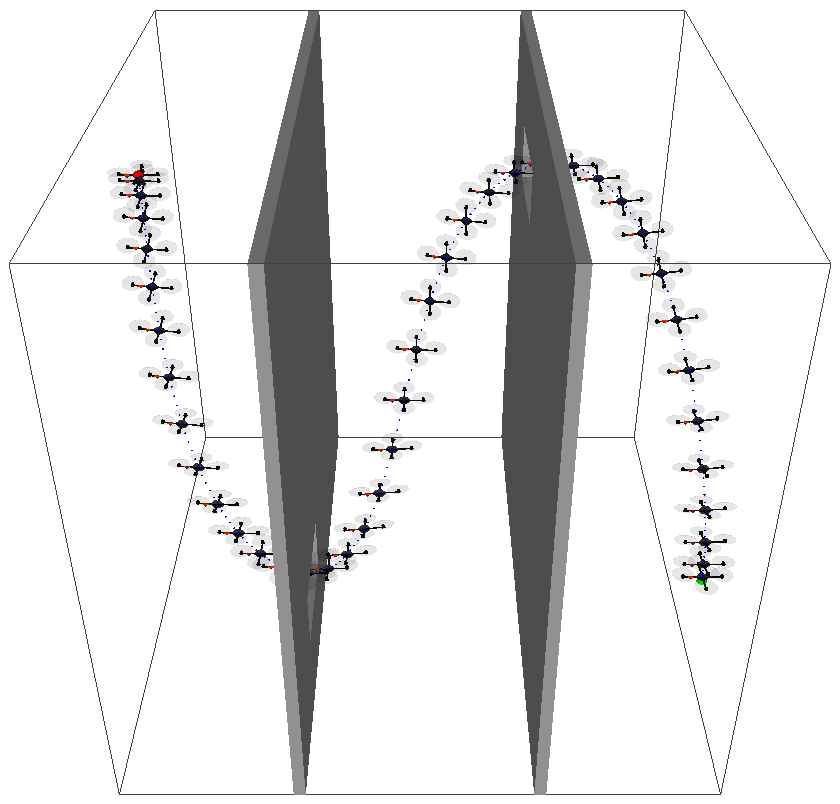}
\vspace{-5pt}
\caption{An asymptotically optimal trajectory computed by our algorithm for a quadrotor helicopter with linearized dynamics in a 10-D state space.}
\label{fig:quadrotorcover}
\vspace{-10pt}
\end{figure}

In this paper, we present Kinodynamic RRT*, an extension of RRT* that overcomes the above limitations by introducing into the algorithm a fixed-final-state-free-final-time controller \cite{lewis95} that exactly and optimally connects any pair of states for any system with controllable linear dynamics in state spaces of arbitrary dimension. Our approach finds asymptotically optimal trajectories in environments with obstacles and bounds on the state and control input, with respect to a cost function that is expressed as a tunable trade-off between the duration of the trajectory and the expended control effort. Moreover, we show that for the rich subclass of systems with a nilpotent dynamics matrix, expressions for optimal connections between pairs of states can be derived in closed-form, and can hence be computed quickly. This means that our algorithm computes asymptotically optimal trajectories for such kinodynamic systems at little additional computational cost compared to RRT* for holonomic robots. Also, our approach can handle non-linear dynamics by linearizing them about the state that is sampled in each iteration of the algorithm.

We note that while we focus our presentation on extending RRT* to kinodynamic systems, also the application of PRM \cite{kavraki96} and path \emph{smoothing} by iterative shortcutting \cite{geraerts07} have thus far been limited to holonomic systems, for these methods too require connecting pairs of states by feasible trajectories. Our approach is equally suited for making PRM and smoothing applicable to robots with differential constraints, and may particularly align well with recent interest in constructing roadmaps containing near-optimal trajectories \cite{marble12}.

We demonstrate the potential of our approach by computing asymptotically optimal trajectories in three challenging motion planning scenarios: (i) a planar robot with a 4-D state space and double integrator dynamics, (ii) an aerial vehicle with a 10-D state space and linearized quadrotor dynamics (see Fig.\ \ref{fig:quadrotorcover}), and (iii) a car-like robot with a 5-D state space and non-linear dynamics.

The remainder of this paper is organized as follows. We begin by discussing related work in Section \ref{sec:relatedwork} and formally defining the problem we discuss in this paper in Section \ref{sec:problemdefinition}. Section \ref{sec:optimalcontrol} describes how an optimal trajectory is computed between any pair of states, and Section \ref{sec:rrt} describes our adapted RRT* algorithm. We describe the extension to non-linear dynamics in Section \ref{sec:nonlinear}, discuss experimental results in Section \ref{sec:results}, and conclude in Section \ref{sec:conclusion}.

\section{Related Work} \label{sec:relatedwork}
The term kinodynamic planning was first introduced in 1993 in \cite{donald93}, which presented a resolution-complete algorithm for optimal planning of robots with discretized double integrator dynamics in low-dimensional workspaces. Kinodynamic planning has since been an active area of research. Incremental sampling-based algorithms, in particular the rapidly-exploring random tree (RRT) approach \cite{lavalle01}, proved to be effective in state spaces of high dimensionality, and is applicable to general dynamics systems as it builds a random tree of trajectories, and complex dynamics can be forward integrated to expand the tree. 

Unfortunately, RRT does not produce optimal trajectories. In fact, the probability that it finds an optimal path is zero \cite{karaman11}. Recently, RRT* was introduced to overcome this problem and guarantees asymptotic optimality \cite{karaman10}; it iteratively builds a tree of trajectories through the state space whose probability of containing an optimal solution approaches 1 as the number of iterations of the algorithm approaches infinity. However, RRT* requires for its \emph{rewiring} step critical to  achieving asymptotic optimality that any pair of states can be \emph{optimally} connected. Therefore, RRT* was introduced for holonomic systems, where any pair of states can be optimally connected by a straight-line trajectory through the state space. 

Several attempts have been made to extend RRT*'s applicability to kinodynamic systems for which a straight-line connection between a pair of states is typically not a valid trajectory due to the system's differential constraints. In \cite{karaman10b}, sufficient conditions were established to ensure asymptotic optimality of the RRT* algorithm for systems with differential constraints, and it was shown how to apply RRT* to two specific instances of kinodynamic systems: the Dubin's car and the double integrator. The approach of \cite{jeon11} generalizes this to arbitrary kinodynamic systems, but has several limitations: to connect pairs of states, it uses the shooting method \cite{burden01} with a constant control input, which is inherently suboptimal. Also, the shooting method can only reach a neighborhood of the final state in general, which requires costly repropagation of the trajectories in the tree descending from such states. Recently, LQR-RRT* was  proposed in \cite{perez12}, and uses an infinite-horizon LQR controller to connect pairs of states. Unfortunately, also this controller will not reach the final state exactly or within a guaranteed neighborhood, even in case of linear dynamics,\footnote{Imagine a system with double integrator dynamics where the state is defined by the robot's position and velocity: an infinite-horizon LQR controller can either reach a state with a specified velocity, or a state with a specified position, but cannot satisfy both as time approaches infinity.} which inherently leads to suboptimality. 

Our approach improves upon this prior work by connecting any pair of states exactly and optimally for systems with controllable linear dynamics, which guarantees that asymptotic optimality is in fact achieved. We accomplish this by extending the well-studied formulation for a fixed final state and fixed final time optimal control problem \cite{lewis95} to derive an optimal, open-loop, fixed final state free final time control policy. A similar approach has been adopted by \cite{tedrake09} for extending RRTs in state space under a dynamic cost-to-go distance metric \cite{glassman10}. In comparison to the latter work, we present a numerical solution that is guaranteed to find a \emph{global optimum} for the general case, and an efficient closed-form solution for the special case of systems with a nilpotent dynamics matrix.

\section{Problem Definition} \label{sec:problemdefinition}
Let $\mathcal{X} = \mathbb{R}^n$ and $\mathcal{U} = \mathbb{R}^m$ be the state space and control input space, respectively, of the robot, and let the dynamics of the robot be defined by the following linear system, which we require to be formally \emph{controllable}:
\begin{align} \label{eq:dynamics}
\dot{\mathbf{x}}[t] & = A\mathbf{x}[t] + B\mathbf{u}[t] + \mathbf{c},
\end{align}
where $\mathbf{x}[t] \in \mathcal{X}$ is the state of the robot, $\mathbf{u}[t] \in \mathcal{U}$ is the control input of the robot, and $A \in \mathbb{R}^{n\times n}$, $B \in \mathbb{R}^{n \times m}$, and $\mathbf{c} \in \mathbb{R}^n$ are constant and given. 

A \emph{trajectory} of the robot is defined by a tuple $\pi = (\mathbf{x}[\,], \mathbf{u}[\,], \tau)$, where $\tau$ is the arrival time or duration of the trajectory, $\mathbf{u} : [0, \tau] \rightarrow \mathcal{U}$ defines the control input along the trajectory, and $\mathbf{x} : [0, \tau] \rightarrow \mathcal{X}$ are the corresponding states along the trajectory given $\mathbf{x}[0]$ with $\dot{\mathbf{x}}[t] = A\mathbf{x}[t] + B\mathbf{u}[t] + \mathbf{c}$.

The \emph{cost} $c[\pi]$ of a trajectory $\pi$ is defined by the function:
\begin{align} \label{eq:cost}
c[\pi] & = \int_0^\tau (1 + \mathbf{u}[t]^T R \mathbf{u}[t])\; \mathrm{d}t,
\end{align}
which penalizes both the duration of the trajectory and the expended control effort, where $R \in \mathbb{R}^{m\times m}$ is positive-definite, constant, and given, and weights the cost of the control inputs relative to each other and to the duration of the trajectory.

Let $\mathcal{X}_\mathrm{free} \subset \mathcal{X}$ define the \emph{free} state space of the robot, which consists of those states that are within user-defined bounds and are collision-free with respect to obstacles in the environment. Similarly, let $\mathcal{U}_\mathrm{free} \subset \mathcal{U}$ define the free control input space of the robot, consisting of control inputs that are within bounds placed on them. This brings us to the formal definition of the problem we discuss in this paper: given a start state $\mathbf{x}_\mathrm{start} \in \mathcal{X}_\mathrm{free}$ and a goal state $\mathbf{x}_\mathrm{goal} \in \mathcal{X}_\mathrm{free}$, find a collision-free trajectory $\pi^*_\mathrm{free}$ between $\mathbf{x}_\mathrm{start}$ and $\mathbf{x}_\mathrm{goal}$ with minimal cost:
\begin{align} \label{eq:problemdefinition}
\pi^*_\mathrm{free} & = \argmin\{\pi\,|\,\mathbf{x}[0] = \mathbf{x}_\mathrm{start} \wedge \mathbf{x}[\tau] = \mathbf{x}_\mathrm{goal} \wedge ~ \nonumber \\ & ~~~~\forall\{t\in[0,\tau]\}\,(\mathbf{x}[t] \in \mathcal{X}_\mathrm{free} \wedge \mathbf{u}[t] \in \mathcal{U}_\mathrm{free})\}\; c[\pi].
\end{align}

We note that the cost function of Eq.\ \eqref{eq:cost} obeys the \emph{optimal substructure property}; let $\pi^*[\mathbf{x}_0, \mathbf{x}_1] = (\mathbf{x}[\,], \mathbf{u}[\,], \tau)$ be the optimal trajectory between $\mathbf{x}_0 \in \mathcal{X}$ and $\mathbf{x}_1 \in \mathcal{X}$, irrespective of bounds and obstacles, and let $c^*[\mathbf{x}_0, \mathbf{x}_1]$ be its cost:
\begin{align}
c^*[\mathbf{x}_0, \mathbf{x}_1] & = \min\{\pi\,|\,\mathbf{x}[0] = \mathbf{x}_0 \wedge \mathbf{x}[\tau] = \mathbf{x}_1\} \; c[\pi]. \label{eq:distance} \\
\pi^*[\mathbf{x}_0, \mathbf{x}_1] & = \argmin\{\pi\,|\,\mathbf{x}[0] = \mathbf{x}_0 \wedge \mathbf{x}[\tau] = \mathbf{x}_1\} \; c[\pi], \label{eq:opttraj}
\end{align}
then for all $0 < t < \tau$ we have: $c^*[\mathbf{x}_0, \mathbf{x}_1] = c^*[\mathbf{x}_0, \mathbf{x}[t]] + c^*[\mathbf{x}[t], \mathbf{x}_1]$. Hence, an optimal collision-free trajectory $\pi^*_\mathrm{free}$ between start and goal  consists of a concatenation of optimal trajectories between a series of successive states $(\mathbf{x}_\mathrm{start}, \mathbf{x}_1, \mathbf{x}_2, \ldots, \mathbf{x}_\mathrm{goal})$ in $\mathcal{X}_\mathrm{free}$. 


\section{Optimally Connecting a Pair of States} \label{sec:optimalcontrol}
A critical component of our approach to solve the problem as defined in Eq.\ \eqref{eq:problemdefinition} is to be able to compute the optimal trajectory $\pi^*[\mathbf{x}_0, \mathbf{x}_1]$ (and its cost $c^*[\mathbf{x}_0, \mathbf{x}_1]$) between any two states $\mathbf{x}_0 \in \mathcal{X}$ and $\mathbf{x}_1 \in \mathcal{X}$, as defined in Eqs.\ \eqref{eq:opttraj} and \eqref{eq:distance}. In this section we discuss how to compute these. It is known from \cite{lewis95} what the optimal control policy is in case a \emph{fixed} arrival time $\tau$ is given, as we review in Section \ref{sec:fixedfinaltime}. We extend this analysis to find the optimal free arrival time in Section \ref{sec:optarrivaltime} and show how to compute the corresponding optimal trajectory in \ref{sec:opttraj}. We discuss practical implementation in Section \ref{sec:pracimpl}.

\subsection{Optimal Control for Fixed Final State and Fixed Final Time} \label{sec:fixedfinaltime}
Given a \emph{fixed} arrival time $\tau$ and two states $\mathbf{x}_0$ and $\mathbf{x}_1$, we want to find a trajectory $(\mathbf{x}[\,]$, $\mathbf{u}[\,], \tau)$ such that $\mathbf{x}[0] = \mathbf{x}_0$, $\mathbf{x}[\tau] = \mathbf{x}_1$, and $\dot{\mathbf{x}}[t] = A\mathbf{x}[t] + B\mathbf{u}[t] + \mathbf{c}$ (for all $0 \leq t \leq \tau$), minimizing the cost function of Eq.\ \eqref{eq:cost}. This is the so-called fixed final state, fixed final time optimal control problem. 

Let $G[t]$ be the \emph{weighted controllability Gramian} given by:
\begin{align} \label{eq:G}
G[t] & = \! \int_0^t \!\exp[A(t - t')] B R^{-1} B^T \exp[A^T (t - t')] \; \mathrm{d}t',
\end{align}
which is the solution to the Lyapunov equation:
\begin{align} \label{eq:dotG}
\dot{G}[t] & = AG[t] + G[t]A^T + B R^{-1} B^T, & G[0] & = 0.
\end{align}
We note that $G[t]$ is a positive-definite matrix for $t > 0$ if the dynamics system of Eq.\ \eqref{eq:dynamics} is controllable.

Further, let $\bar{\mathbf{x}}[t]$ describe what the state $\mathbf{x}$ (starting in $\mathbf{x}_0$ at time 0) would be at time $t$ if no control input were applied:
\begin{align} \label{eq:barx}
\bar{\mathbf{x}}[t] & = \exp[At]\mathbf{x}_0 + \int_0^t \exp[A(t - t')]\mathbf{c}\; \mathrm{d}t',
\end{align}
which is the solution to the differential equation:
\begin{align} \label{eq:dotbarx}
\dot{\bar{\mathbf{x}}}[t] & = A\bar{\mathbf{x}}[t] + \mathbf{c}, & \bar{\mathbf{x}}[0] & = \mathbf{x}_0.
\end{align}

Then, the optimal control policy for the fixed final state, fixed final time optimal control problem is given by:
\begin{align} \label{eq:controlpolicy}
\mathbf{u}[t] = R^{-1}B^T \exp[A^T(\tau - t)]G[\tau]^{-1}(\mathbf{x}_1 - \bar{\mathbf{x}}[\tau]),
\end{align}
which is an \emph{open-loop} control policy. We refer the reader to \cite{lewis95} for details on the derivation of this equation.

\subsection{Finding the Optimal Arrival Time} \label{sec:optarrivaltime}
To find an optimal trajectory $\pi^*[\mathbf{x}_0, \mathbf{x}_1]$ between $\mathbf{x}_0$ and $\mathbf{x}_1$ as defined by Eq.\ \eqref{eq:opttraj}, we extend the above analysis to solve the fixed final state, \emph{free} final time optimal control problem, in which we can choose the arrival time $\tau$ freely to minimize the cost function of Eq.\ \eqref{eq:cost}. 

To find the optimal arrival time $\tau^*$, we proceed as follows. By filling in the control policy of Eq.\ \eqref{eq:controlpolicy} into the cost function of Eq.\ \eqref{eq:cost} and evaluating the integral, we find a closed-form expression for the cost of the optimal trajectory between $\mathbf{x}_0$ and $\mathbf{x}_1$ for a given (fixed) arrival time $\tau$:
\begin{align}
c[\tau] & = \tau + (\mathbf{x}_1 - \bar{\mathbf{x}}[\tau])^T G[\tau]^{-1} (\mathbf{x}_1 - \bar{\mathbf{x}}[\tau]).
\end{align}

The optimal arrival time $\tau^*$ is the value of $\tau$ for which this function is minimal:
\begin{align}
\tau^* = \argmin\{\tau > 0\} \; c[\tau],
\end{align}
and the cost of the optimal trajectory between $\mathbf{x}_0$ and $\mathbf{x}_1$ as defined in Eq.\ \eqref{eq:distance} is given by $c^*[\mathbf{x}_0, \mathbf{x}_1] = c[\tau^*]$.

\begin{figure}
\centering
\includegraphics[width=0.6\linewidth]{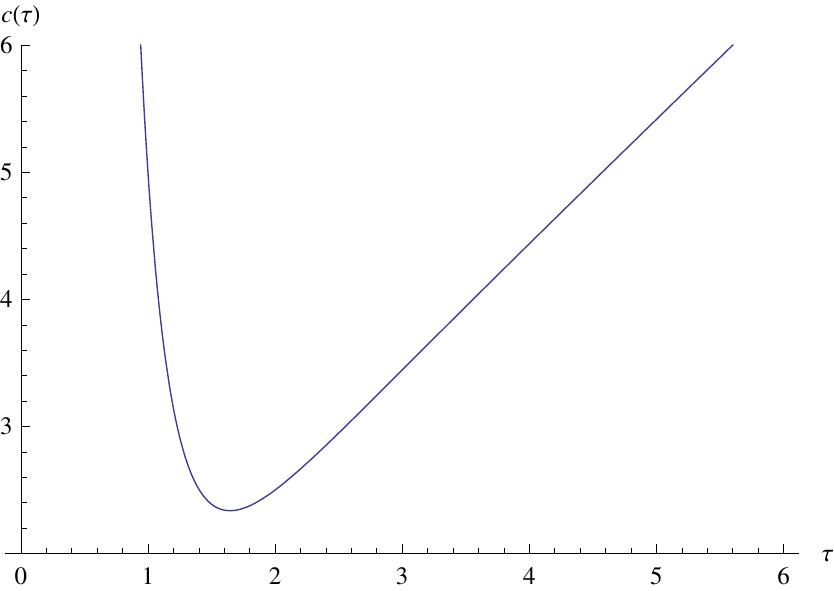}
\caption{Plot of the function $c[\tau]$ for a robot moving on the 1-D line with double integrator dynamics ($A = ((0, 1), (0, 0))$, $B = (0, 1)^T$, $\mathbf{c} = \mathbf{0}$, $R = 1$) between $\mathbf{x}_0 = (0,0)^T$ and $\mathbf{x}_1 = (1,1)^T$. The optimal arrival time is $\tau^* = \sqrt{7} - 1 \approx 1.65$, which is where the function $c[\tau]$ is minimal.}
\vspace{-10pt}
\label{fig:costfunction}
\end{figure}

The optimal arrival time $\tau^*$ is found by taking the derivative of $c[\tau]$ with respect to $\tau$ (which we denote $\dot{c}[\tau]$), and solving $\dot{c}[\tau] = 0$ for $\tau$. The derivative is given by:
\begin{align} \label{eq:dotc}
\dot{c}[\tau] & = 1 - 2(A\mathbf{x}_1 + \mathbf{c})^T \mathbf{d}[\tau] - \mathbf{d}[\tau]^T BR^{-1}B^T\mathbf{d}[\tau], \end{align}
where we define:
\begin{align}
\mathbf{d}[\tau] = G[\tau]^{-1}(\mathbf{x}_1 - \bar{\mathbf{x}}[\tau]).
\end{align}
It should be noted that the function $c[\tau]$ may have multiple local minima. Also, note that $c[\tau] > \tau$ for all $\tau > 0$, since $G[\tau]$ is positive-definite (see Fig.\ \ref{fig:costfunction}).

\subsection{Computing the Optimal Trajectory} \label{sec:opttraj}
Given the optimal arrival time $\tau^*$ as defined above, we find the corresponding optimal trajectory $\pi^*[\mathbf{x}_0, \mathbf{x}_1] = (\mathbf{x}[\,],\mathbf{u}[\,], \tau^*)$, as defined in Eq.\ \eqref{eq:opttraj}, as follows. Let us define:
\begin{align} \label{eq:y}
\mathbf{y}[t] = \exp[A^T(\tau^* - t)]\mathbf{d}[\tau^*],
\end{align}
such that the optimal control policy (see Eq.\ \eqref{eq:controlpolicy}) is given by:
\begin{align} \label{eq:u}
\mathbf{u}[t] = R^{-1}B^T\mathbf{y}[t].
\end{align}
Filling in this optimal control policy into Eq.\ \eqref{eq:dynamics} gives us the differential equation for the state $\mathbf{x}[\,]$:
\begin{align} \label{eq:dotx}
\dot{\mathbf{x}}[t] & = A\mathbf{x}[t] + BR^{-1}B^T\mathbf{y}[t] + \mathbf{c}, & \mathbf{x}[\tau^*] & = \mathbf{x}_1.
\end{align}
Noting that Eq.\ \eqref{eq:y} is the solution to the differential equation:
\begin{align} \label{eq:doty}
\dot{\mathbf{y}}[t] & = -A^T\mathbf{y}[t], & \mathbf{y}[\tau^*] = \mathbf{d}[\tau^*],
\end{align}
and combining this with Eq.\ \eqref{eq:dotx} gives us the composite differential equation:
\begin{align}
\begin{bmatrix}\dot{\mathbf{x}}[t] \\ \dot{\mathbf{y}}[t] \end{bmatrix} & \! = \! \begin{bmatrix} A & BR^{-1}B^T \\ 0 & -A^T \end{bmatrix} \begin{bmatrix}\mathbf{x}[t]\\ \mathbf{y}[t]\end{bmatrix} + \begin{bmatrix} \mathbf{c} \\ \mathbf{0} \end{bmatrix}\!, & \begin{bmatrix} \mathbf{x}[\tau^*] \\ \mathbf{y}[\tau^*] \end{bmatrix} & \! = \! \begin{bmatrix} \mathbf{x}_1 \\ \mathbf{d}[\tau^*] \end{bmatrix}\!, \label{eq:trajdiff}
\end{align}
which has as solution:
\begin{align}
\begin{bmatrix}\mathbf{x}[t] \\ \mathbf{y}[t] \end{bmatrix} &  = \exp\!\bigg[ \begin{bmatrix} A & BR^{-1}B^T \\ 0 & -A^T \end{bmatrix}(t-\tau^*)\bigg] \begin{bmatrix}\mathbf{x}_1 \\ \mathbf{d}[\tau^*]\end{bmatrix} + ~ \nonumber \\
& ~~~ \int_{\tau^*}^t \exp\!\bigg[\begin{bmatrix} A & BR^{-1}B^T \\ 0 & -A^T \end{bmatrix}(t - t') \bigg] \begin{bmatrix} \mathbf{c} \\ \mathbf{0} \end{bmatrix} \mathrm{d}t'. \label{eq:traj}
\end{align}
This gives us $\mathbf{x}[t]$ and, using Eq.\ \eqref{eq:u}, $\mathbf{u}[t]$ for all $0 < t < \tau^*$, which completely determines $\pi^*[\mathbf{x}_0, \mathbf{x}_1]$.

\subsection{Practical Implementation} \label{sec:pracimpl}
For the implementation of the above computations in practice, we distinguish the special case in which matrix $A$ is \emph{nilpotent}, in which case we can derive a closed-form solution for the optimal trajectory, from the general case, in which case we can find the optimal trajectory numerically.

If matrix $A \in \mathbb{R}^{n\times n}$ is \emph{nilpotent}, i.e. $A^n = 0$, which is not uncommon as we will see in Section \ref{sec:results}, $\exp[At]$ has a closed-form expression in the form of an $(n-1)$-degree matrix polynomial in $t$. As a result, the integrals of Eqs.\ \eqref{eq:G} and \eqref{eq:barx} can be evaluated exactly to obtain closed-form expressions for $G[\tau]$ and $\bar{\mathbf{x}}[\tau]$. Solving $\dot{c}[\tau] = 0$ for $\tau$ to find the optimal arrival time $\tau^*$ then amounts to finding the roots of a (high-degree) polynomial in $\tau$. Various methods exist to find all roots of a polynomial \cite{burden01}, which gives us the global minimum of $c[\tau]$ and the corresponding optimal arrival time $\tau^*$. Subsequently, the nilpotence of $A$ implies that the matrix $\big[\begin{smallmatrix} A & BR^{-1}B^T \\ 0 & -A^T \end{smallmatrix}\big]$ is nilpotent as well, which means that Eq.\ \eqref{eq:traj} can be evaluated exactly to obtain a closed form expression for the optimal trajectory (states and control inputs) between any two states $\mathbf{x}_0$ and $\mathbf{x}_1$. 

For a general (not nilpotent) matrix $A$, we integrate $\dot{G}[t]$ and $\dot{\bar{\mathbf{x}}}[t]$ forward in time according to Eqs.\ \eqref{eq:dotG} and \eqref{eq:dotbarx} using the 4th-order Runge-Kutta method \cite{burden01}, which gives us $G[\tau]$, $\bar{\mathbf{x}}[\tau]$, and $c[\tau]$ for increasing $\tau > 0$. We keep track of the minimal cost $c^* = c[\tau]$ we have seen so far and the corresponding arrival time, as we perform the forward integration for increasing $\tau > 0$. Since $c[\tau] > \tau$ for all $\tau > 0$, it suffices to terminate the forward integration at $\tau = c^*$ to guarantee that a global minimum $c^*$ of $c[\tau]$, and the corresponding optimal arrival time $\tau^*$, has been found. This procedure also gives us $\mathbf{d}[\tau^*]$, which we use to subsequently reconstruct the optimal trajectory between $\mathbf{x}_0$ and $\mathbf{x}_1$, by integrating the differential equation \eqref{eq:trajdiff} backward in time for $\tau^* > t > 0$ using 4th-order Runge-Kutta.

\section{Kinodynamic RRT*} \label{sec:rrt}
To find the optimal collision-free trajectory $\pi^*_\mathrm{free}$ as defined in Eq.\ \eqref{eq:problemdefinition}, given the ability to find an optimal trajectory between any pair of states as described above, we use an adapted version of RRT*, since RRT* is known to achieve \emph{asymptotic optimality}; that is, as the number of iterations of the algorithm approaches infinity, the probability that an optimal path has been found approaches 1. The algorithm is given in Fig.\ \ref{fig:algorithm}. 

The algorithm builds a tree $\mathcal{T}$ of trajectories in the free state space rooted in the start state. In each iteration $i$ of the algorithm, a state $\mathbf{x}_i$ is sampled from the free state space $\mathcal{X}_\mathrm{free}$ (line \ref{line:sample}) to become a new node of the tree (line \ref{line:add}). For each new node a parent is found among neighboring nodes already in the tree, i.e. the nodes $\mathbf{x}$ for which $c^*[\mathbf{x}, \mathbf{x}_i] < r$ for some neighbor radius $r$. The node $\mathbf{x}$ that is chosen as parent is the node for which the optimal trajectory $\pi[\mathbf{x}, \mathbf{x}_i]$ to the new node is collision-free (i.e. the states and control inputs along the trajectory are in the respective free spaces) and results in a minimal cost between the root node ($\mathbf{x}_\mathrm{start}$) and the new node (lines \ref{line:parent}-\ref{line:cost}). Subsequently, it is attempted to decrease the cost from the start to other nodes in the tree by connecting the new node to neighboring nodes in the tree, i.e. the nodes $\mathbf{x}$ for which $c^*[\mathbf{x}_i, \mathbf{x}] < r$. For each state $\mathbf{x}$ for which the connection is collision-free and results in a lower cost to reach $\mathbf{x}$ from the start, the new node $\mathbf{x}_i$ is made the parent of $\mathbf{x}$ (lines \ref{line:rewire}-\ref{line:newparent}). Then, the algorithm continues with a new iteration. If this is repeated indefinitely, an optimal path between $\mathbf{x}_\mathrm{start}$ and $\mathbf{x}_\mathrm{goal}$ will emerge in the tree.

\begin{figure}[t]
\textsc{KinodynamicRRT*}$[\mathbf{x}_\mathrm{start} \in \mathcal{X}_\mathrm{free}, \mathbf{x}_\mathrm{goal} \in \mathcal{X}_\mathrm{free}]$

\begin{algorithmic}[1]
  \STATE $\mathcal{T} \leftarrow \{\mathbf{x}_\mathrm{start}\}$.
  \FOR {$i \in [1, \infty)$}
    \STATE Randomly sample $\mathbf{x}_i \in \mathcal{X}_\mathrm{free}$. \label{line:sample}
    \STATE $\mathbf{x} \leftarrow \argmin\{\mathbf{x} \in \mathcal{T}\,|\,c^*[\mathbf{x}, \mathbf{x}_i] < r \wedge~$ \\ ~~~~~~\textsc{CollisionFree}$[\pi^*[\mathbf{x}, \mathbf{x}_i]]\}\,(\mathrm{cost}[\mathbf{x}] + c^*[\mathbf{x}, \mathbf{x}_i])$. \label{line:parent}
    \STATE $\mathrm{parent}[\mathbf{x}_i] \leftarrow \mathbf{x}$.
    \STATE $\mathrm{cost}[\mathbf{x}_i] \leftarrow \mathrm{cost}[\mathbf{x}] + c^*[\mathbf{x}, \mathbf{x}_i]$. \label{line:cost}
      \FORALL {$\{\mathbf{x} \in \mathcal{T} \cup \{\mathbf{x}_\mathrm{goal}\} \,|\, c^*[\mathbf{x}_i, \mathbf{x}] < r \wedge \mathrm{cost}[\mathbf{x}_i] + c^*[\mathbf{x}_i, \mathbf{x}] < \mathrm{cost}[\mathbf{x}] \wedge~\!\!$\textsc{CollisionFree}$[\pi^*[\mathbf{x}_i, \mathbf{x}]]\}$} \label{line:rewire}
        \STATE $\mathrm{cost}[\mathbf{x}] \leftarrow \mathrm{cost}[\mathbf{x}_i] + c^*[\mathbf{x}_i, \mathbf{x}]$.
        \STATE $\mathrm{parent}[\mathbf{x}] \leftarrow \mathbf{x}_i$. \label{line:newparent}
      \ENDFOR
      \STATE $\mathcal{T} \leftarrow \mathcal{T} \cup \{\mathbf{x}_i\}$. \label{line:add}
  \ENDFOR
\end{algorithmic}
\vspace{-5pt}
\caption{The adapted RRT* algorithm. The tree $\mathcal{T}$ is represented as a set of states. Each state $\mathbf{x}$ in the tree has two attributes: a pointer $\mathrm{parent}[\mathbf{x}]$ to its parent state in the tree, and a number $\mathrm{cost}[\mathbf{x}]$ which stores the cost of the trajectory in the tree between the start state and $\mathbf{x}$. Further, we define \textsc{CollisionFree}$[\mathbf{x}[\,], \mathbf{u}[\,], \tau] = \forall\{t \in [0,\tau]\}\,(\mathbf{x}[t] \in \mathcal{X}_\mathrm{free} \wedge \mathbf{u}[t] \in \mathcal{U}_\mathrm{free})$.} \label{fig:algorithm}
\vspace{-10pt}
\end{figure}

The algorithm as given in Fig.\ \ref{fig:algorithm} differs subtly from the standard RRT* algorithm. First of all, we have defined our problem as finding a trajectory that exactly arrives at a goal state, rather than a goal region as is common in RRT*. As a consequence, we explicitly add the goal state to the set of states that is considered for a forward connection from a newly sampled node in line \ref{line:rewire}, even if the goal is not (yet) part of tree. Also, typical RRT* implementations include a ``steer'' module, which lets the tree grow \emph{towards} a sampled state (but not necessarily all the way), and adds the endpoint of a partial trajectory as node to the tree \cite{karaman11}. Since it is non-trivial given our formulation to compute a partial trajectory of a specified maximum cost, our algorithm attempts a full connection to the sampled state, and adds the sampled state itself as a node to the tree. These changes do not affect the asymptotic optimality guarantee of the algorithm.

In the original RRT* algorithm, the neighbor radius $r$ can be decreased over the course of the algorithm as a function $r = ((\gamma/\zeta_d)\log[i]/i)^{1/d}$ of the number of nodes $i$ currently in the tree, without affecting the asymptotic optimality guarantee, where $d$ is the dimension of the state space, $\zeta_d$ is the volume of a $d$-dimensional unit ball,  $\gamma > 2^d(1+1/d)\mu[\mathcal{X}_\mathrm{free}]$, and $\mu[\mathcal{X}_\mathrm{free}]$ is the volume of the state space  \cite{karaman10}. For non-Euclidean distance measures, such as our function $c^*[\mathbf{x}_0, \mathbf{x}_1]$, let $\mathcal{R}[\mathbf{x}, r]$ be the set of states that can reach $\mathbf{x}$ or are reachable from $\mathbf{x}$ with cost less than $r$:
\begin{align}
\mathcal{R}[\mathbf{x}, r] = \{ \mathbf{x}' \in \mathcal{X}\,|\, c^*[\mathbf{x}, \mathbf{x}'] < r \vee c^*[\mathbf{x}', \mathbf{x}] < r\}.
\end{align}
Then the neighbor radius $r$ must be set such that a ball of volume $\gamma \log[i]/i$ is contained within $\mathcal{R}[\mathbf{x}, r]$ \cite{karaman10b}. A finite radius $r$ always exists in our case such that this holds, since we  require that the system's dynamics are formally controllable.


\section{Systems with Non-Linear Dynamics} \label{sec:nonlinear}
We have presented our algorithm for linear dynamics systems of the type of Eq.\ \eqref{eq:dynamics}, but we can apply our algorithm to non-linear dynamics as well through linearization. Let the non-linear dynamics of the robot be defined by a function $\mathbf{f}$:
\begin{align}
\dot{\mathbf{x}}[t] = \mathbf{f}[\mathbf{x}[t], \mathbf{u}[t]].
\end{align}
We can locally approximate the dynamics by linearizing the function $\mathbf{f}$ to obtain a system of the form of Eq.\ \eqref{eq:dynamics}, with:
\begin{align}
A & = \frac{\partial \mathbf{f}}{\partial \mathbf{x}}[\hat{\mathbf{x}}, \hat{\mathbf{u}}], & B & = \frac{\partial \mathbf{f}}{\partial \mathbf{u}}[\hat{\mathbf{x}}, \hat{\mathbf{u}}], & \mathbf{c} & = \mathbf{f}[\hat{\mathbf{x}}, \hat{\mathbf{u}}] - A\hat{\mathbf{x}} - B\hat{\mathbf{u}},
\end{align}
where $\hat{\mathbf{x}}$ is the state and $\hat{\mathbf{u}}$ is the control input about which the dynamics are linearized. The resulting linear system is a first-order Taylor approximation of the non-linear dynamics, and is approximately valid only in the vicinity of $\hat{\mathbf{x}}$ and $\hat{\mathbf{u}}$. 

We adapt the algorithm of Fig.\ \ref{fig:algorithm} to non-linear dynamics by (re)linearizing $\mathbf{f}$ in each iteration of the algorithm about $\hat{\mathbf{x}} = \mathbf{x}_i$ and $\hat{\mathbf{u}} = \mathbf{0}$ after a new state $\mathbf{x}_i$ is sampled in line \ref{line:sample}. The resulting linear dynamics are then used in the subsequent computations of the functions $c^*$ and $\pi^*$ (note that this only works if the linearized dynamics are controllable). We choose $\hat{\mathbf{x}} = \mathbf{x}_i$ since it is either the start or the end point of any trajectory computed in that iteration of the algorithm, and we choose $\hat{\mathbf{u}} = \mathbf{0}$ since the cost function (see Eq.\ \eqref{eq:cost}) explicitly penalizes deviations of the control input from zero. 

The linearization is only a valid approximation if the computed trajectories do not venture too much away from the linearization point. As over the course of the algorithm the distances between states get shorter (due to a decreasing neighbor radius $r$) and the trajectories in the tree get more optimal (hence having control inputs closer to zero), this approximation becomes increasingly more reasonable. It can be helpful, though, to let the neighbor radius $r$ not exceed a certain maximum within which the linearizations can be assumed valid.

\section{Experimental Results} \label{sec:results}
We experimented with our implementation on three kinodynamic systems; a double integrator disk robot operating in the plane, a quadrotor robot operating in three space, and a non-holonomic car-like robot operating in the plane, which are discussed in detail in Sections \ref{sec:didetails}, \ref{sec:qdetails}, and \ref{sec:nhdetails}, respectively. Simulation results are subsequently analyzed in Section \ref{sec:analysis}.

\subsection{Linear Double Integrator Model} \label{sec:didetails}
The double integrator robot is a circular robot capable of moving in any direction by controlling its acceleration. Its state space is four-dimensional, and its linear dynamics are described by:
\begin{align}
\bf{x} &  = \begin{bmatrix} \bf{p} \\ \bf{v} \end{bmatrix}, & A & = \begin{bmatrix} 0 & I \\ 0 & 0 \end{bmatrix}, & B & = \begin{bmatrix} 0 \\ I \end{bmatrix}, & R & = rI,
\end{align}
$\bf{u} = \bf{a}$, and $\bf{c} = \bf{0}$, where $\bf{p}$ describes its position in the plane, $\bf{v}$ its velocity, and $\bf{a}$ its acceleration. Further, we set bounds such that $\mathbf{p} \in [0, 200]\times[0,100]$ (m), $\mathbf{v} \in [-10, 10]^2$ (m/s), and $\mathbf{u} = \mathbf{a} \in [-10, 10]^2$ (m/s$^2$). The control penalty $r$ was set to 0.25 as this permitted the robot to reach velocities near its bounds but not frequently exceed them. 

\begin{figure}
\centering
\includegraphics[width=0.85\linewidth]{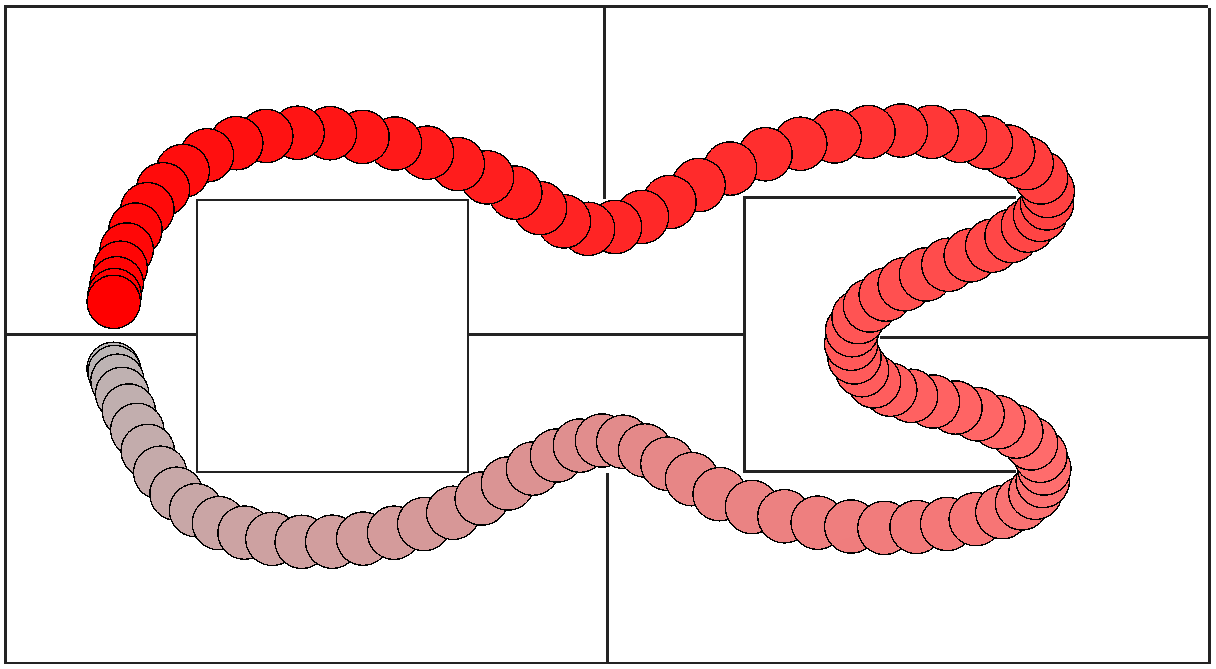}
\caption{Asymptotically optimal trajectory for double integrator robot after 100,000 nodes were added to the tree.}
\vspace{-10pt}
\label{fig:doubleintegrator}
\end{figure}

We experimented with this model in the environment of Fig.\ \ref{fig:doubleintegrator}. Clearly, $A$ is nilpotent as $A^2 = 0$, so we can use both the closed-form and the numerical method for computing connections between states. 

\subsection{Linearized Quadrotor Model} \label{sec:qdetails}
The quadrotor helicopter was modeled after the Ascending Technologies' ResearchPilot. Its state $\mathbf{x} = (\mathbf{p}^T, \mathbf{v}^T, \mathbf{r}^T, \mathbf{w}^T)^T$ is 12-dimensional, consisting of three-dimensional position $\mathbf{p}$, velocity $\mathbf{v}$, orientation $\mathbf{r}$ (rotation about axis $\bf{r}$ by angle $||\bf{r}||$), and angular velocity $\mathbf{w}$. Its dynamics are non-linear \cite{michael10}, but are well-linearizable about the hover point of the quadrotor.  The linearization is (very) sensitive though to deviations in the yaw. Fortunately, the yaw is a redundant degree of freedom, so in our linearization, we constrain the yaw (and its derivative) to zero. This gives a reduced ten-dimensional state and three-dimensional control input, with the following linearized dynamics:
\begin{align}
\bf{x} & = \begin{bmatrix} \bf{p} \\ \bf{v} \\ \bf{r} \\ \bf{w} \end{bmatrix}, & \mathbf{u} & = \begin{bmatrix} u_f \\ u_x \\ u_y \end{bmatrix}, & A & = \begin{bmatrix} 0 & I & 0 & 0 \\
                    0 & 0 & \left[\begin{smallmatrix} 0 & g \\ -g & 0 \\ 0 & 0 \end{smallmatrix}\right] & 0 \\
                    0 & 0 & 0 & I \\
                    0 & 0 & 0 & 0 \end{bmatrix}\!,
\end{align}
\begin{align}
B & = \begin{bmatrix} \mathbf{0} & 0 \\
                      \left[\begin{smallmatrix} 0 \\ 0 \\ 1/m \end{smallmatrix}\right] & 0 \\
                      \mathbf{0} & 0 \\
                      \mathbf{0} & \ell I/j \end{bmatrix}\!, & \mathbf{c} & = \mathbf{0}, & R & = \begin{bmatrix} \frac{1}{4} & 0 & 0 \\
                                                                                     0 & \frac{1}{2} & 0 \\
                                                                                     0 & 0 & \frac{1}{2} \end{bmatrix},
\end{align}
where $\mathbf{r}$ and $\mathbf{w}$ are two-dimensional (with their third component implicitly zero), $g = 9.8$m/s$^2$ is the gravity, $m$ is the mass of the quadrotor (kg), $\ell$ the distance between the center of the vehicle and each of the rotors (m), and $j$ is the moment of inertia of the vehicle about the axes coplanar with the rotors (kg m$^2$). The control input $\bf{u}$ consists of three components: $u_f$ is the total thrust of the rotors relative to the thrust needed for hovering, and $u_x$ and $u_y$ describe the relative thrust of the rotors producing roll and pitch, respectively. Further, the bounds are defined as $\mathbf{p} \in [0, 5]^3$ (m), $\mathbf{v} \in [-5, 5]^3$ (m/s), $\mathbf{r} \in [-1, 1]^2$ (rad), $\mathbf{w} \in [-5, 5]^2$ (rad/s), $u_f \in [-4.545, 9.935]$ (N), and $u_x, u_y \in [-3.62, 3.62]$ (N). The matrix $R$ was chosen such that producing force is penalized equally for each rotor.

\begin{figure}
\centering
\includegraphics[width=0.85\linewidth]{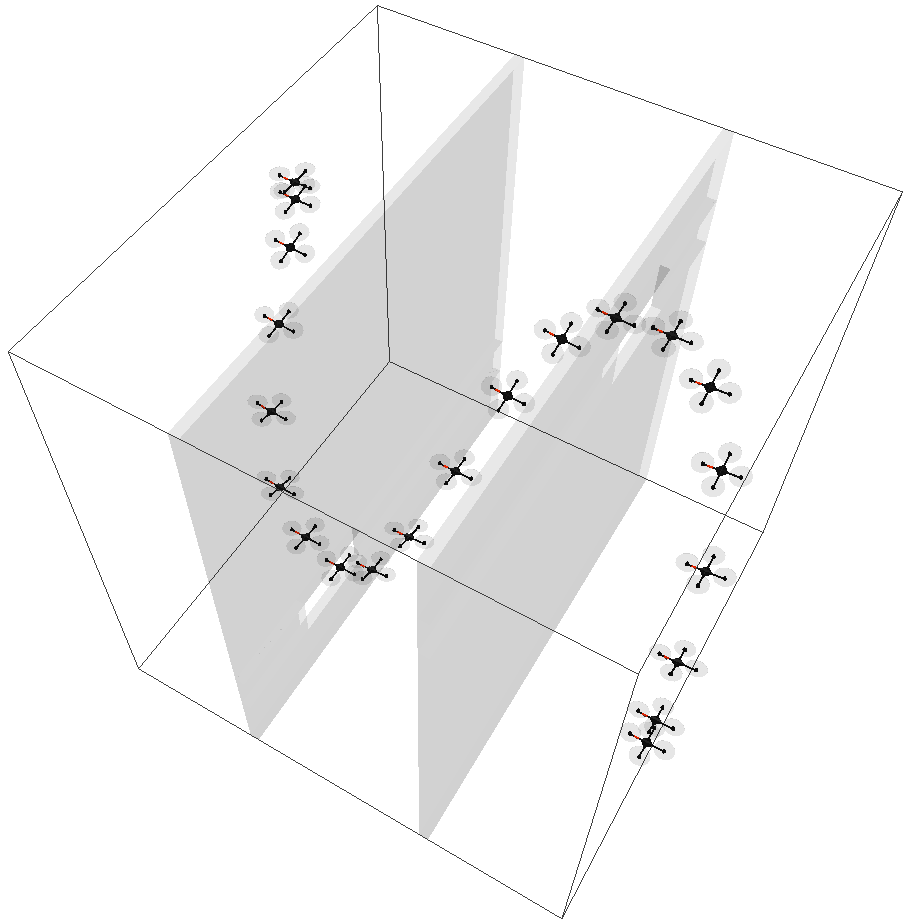}
\caption{Asymptotically optimal trajectory for quadrotor robot after 60,000 nodes were added to the tree.}
\vspace{-10pt}
\label{fig:quadrotor}
\end{figure}

The quadrotor simulations were performed in the environment of Fig.\ \ref{fig:quadrotor} for the linearized dynamics. Clearly, $A$ is nilpotent as it is strictly upper diagonal, so we can use both the closed-form and the numerical method for computing connections between states. 

\subsection{Non-Linear Car-Like Model} \label{sec:nhdetails}
The car-like robot has a five-dimensional state $\mathbf{x} = (x, y, \theta, v, \kappa)^T$, consisting of its planar position $(x,y)$ (m), its orientation $\theta$ (rad), speed $v$ (m/s), and curvature $\kappa$ (m$^{-1}$). The control input $\mathbf{u} = (u_v, u_\kappa)^T$ is two-dimensional and consists of the derivatives of speed and curvature, respectively. The dynamics are non-linear, and described by $\dot{\mathbf{x}} = \mathbf{f}[\mathbf{x}, \mathbf{u}]$, with $\mathbf{f}$ given by:
\begin{align}
\dot{x} & = v \cos \theta, & \dot{y} & = v \sin \theta, & \dot{\theta} & = v \kappa, & \dot{v} & = u_v, & \dot{\kappa} & = u_\kappa.
\end{align}
For this system, we repeatedly linearize the dynamics about the last sampled state, as described in Section \ref{sec:nonlinear}. Let this state be $\hat{\mathbf{x}} = (\hat{x}, \hat{y}, \hat{\theta}, \hat{v}, \hat{\kappa})^T$, then the linearized dynamics are given by:
\begin{align}
A & = \begin{bmatrix} 0 & 0 & -\hat{v} \sin \hat{\theta} & \cos \hat{\theta} & 0 \\
                      0 & 0 & \hat{v} \cos \hat{\theta} & \sin \hat{\theta} & 0 \\
                      0 & 0 & 0 & \hat{\kappa} & \hat{v} \\
                      0 & 0 & 0 & 0 & 0\\
                      0 & 0 & 0 & 0 & 0 \end{bmatrix}\!, & B & = \begin{bmatrix} 0 & 0 \\ 0 & 0 \\ 0 & 0 \\ 1 & 0 \\ 0 & 1 \end{bmatrix}\!,
\end{align}
and $\mathbf{c} = \mathbf{f}[\hat{\mathbf{x}}, \mathbf{0}] - A\hat{\mathbf{x}}$, where bounds were set such that $p_x \in [0, 200]$, $p_y \in [0, 100]$, $\theta \in [-\pi, \pi]$, $v \in (0, 10]$, $\kappa \in [-0.25, 0.25]$. We note that the velocity must be non-zero, otherwise the resulting linear dynamics are not controllable.

\begin{figure}
\centering
\includegraphics[width=0.85\linewidth]{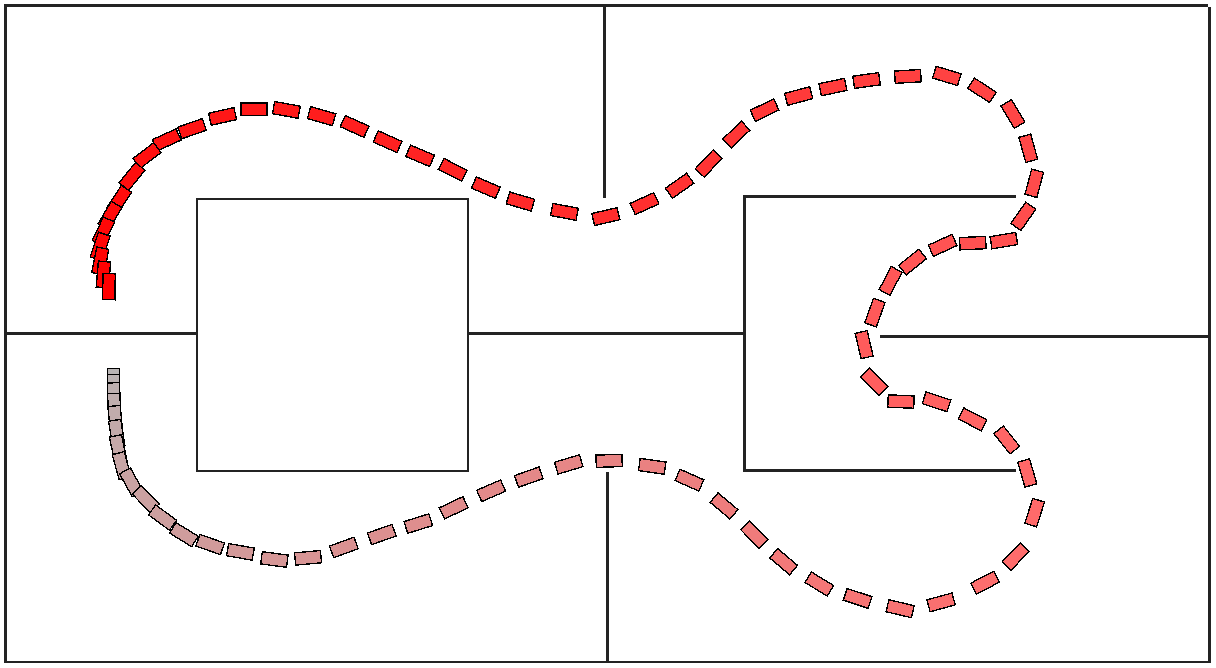}
\caption{Asymptotically optimal trajectory for the car-like robot after 60,000 nodes were added to the tree.}
\label{fig:nonholonomic}
\end{figure}

The car-like robot experiments were performed in the environment of Fig.\ \ref{fig:nonholonomic}, the same environment as the double integrator. Also in this case, the dynamics matrix $A$ is nilpotent for all linearizations, so we can use both the closed-form and the numerical method for computing connections between states. 

\begin{table}
\centering
\caption{Timing data for the first 5000 nodes of each simulation (s).}
\begin{tabular}{l|ccc|ccc}
\hline
 & \multicolumn{3}{c|}{Closed Form} & \multicolumn{3}{c}{Runge Kutta 4} \\ 
nodes & DblInt & QuadRtr & Car & DblInt & QuadRtr & Car \\ \hline
1000 & 19.75  & 116.2 & 5.416  & 969.8 & 4644 & 274.8 \\ 
2000 & 43.26   & 274.9 & 12.43 & 3638 & 10819 & 331.2 \\ 
3000 & 72.59  & 507.3 & 21.74 & 7872 & 20284 & 405.6 \\ 
4000 & 108.6 & 841.6  & 31.99 & 13479 & 33401 & 497.4 \\ 
5000 & 150.4 & 1168 & 42.94 & 20772 & 68703 & 606.3 \\ \hline
\end{tabular}
\label{tab:timings}
\vspace{-10pt}
\end{table}

\subsection{Analysis of Results} \label{sec:analysis}
We used our algorithm to compute asymptotically optimal trajectories for the double integrator, the quadrotor, and the car-like robots. They are shown in Figs.\ \ref{fig:doubleintegrator}, \ref{fig:quadrotor}, and \ref{fig:nonholonomic}. While the paths found for the double integrator and quadrotor robots appear continuous and smooth, in the case of the car-like robot effects of linearization are clearly visible; the robot appears to skid sideways to some extent, as if drifting through the curves.

Table \ref{tab:timings} shows the time required to expand the first 5,000 nodes for all three systems using both the closed form method and the numerical 4th-order Runge-Kutta (RK4) method for computing connections between states. It is clear that the closed-form method executed much more quickly than the RK4 method in all cases. On average, we see a factor of 45 (!) difference in running time. Using either of the two methods resulted in solutions with comparable costs after expanding the same number of nodes; variations that occurred were a result of numerical errors. 

We also see that nodes were less quickly processed for the double integrator than for the car-like robot, despite a lower dimensionality. This is because for the double integrator and quadrotor experiments, we used a neighbor radius of $r = \infty$, while in the case of the non-holonomic car-like robot only connections to states within a tight radius (approximately corresponding to connections within one width of the road) were accepted. This radius was imposed on the car-like system to ensure short connections as the linearization breaks down over large distances, but it also demonstrates the positive effect of using a reduced radius on performance. Processing nodes for the quadrotor experiment appeared most computationally intensive. This is a result of the high-dimension of its state space. The numbers of Table \ref{tab:timings} also highlight the quadratic nature of the algorithm: the total accumulated running time is a quadratic function of the number of nodes that have been added to the tree.

\begin{figure}
\centering
\includegraphics[width=1.0\linewidth]{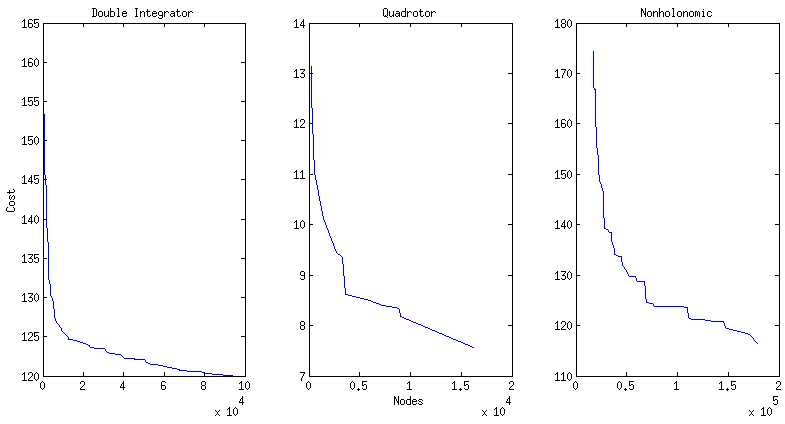}
\vspace{-10pt}
\caption{Graphs showing the cost of the current-best solution as a function of the number of nodes in the tree for (from left to right) the double integrator, the quadrotor, and the car-like robot experiments.}
\vspace{-10pt}
\label{fig:graphs}
\end{figure}

Figure \ref{fig:graphs} shows the cost of the current-best solution for each experiment as more nodes are added to the tree. A total of 100,000 nodes were expanded in the double integrator robot simulation, 60,000 nodes in the quadrotor robot simulation, and 200,000 nodes in the non-holonomic robot simulation. In all cases we see that a high cost solution is found in relatively few nodes, and that these solutions are quickly refined as a result of the RRT* rewiring procedure. Necessarily these refinements plain off as the solutions approach the asymptotic optimum.

\section{Discussion, Conclusion, and Future Work} \label{sec:conclusion}
We have presented Kinodynamic RRT*, an incremental sampling-based approach that extends RRT* for asymptotically optimal motion planning for robots with differential constraints. Our approach achieves asymptotically optimality by using a fixed-final-state-free-final-time optimal control formulation that connects any pair of states exactly and optimally for systems with controllable linear dynamics. We have shown that tor the rich subclass of systems with a nilpotent dynamics matrix, such trajectories can be computed efficiently, making asymptotically optimal planning computationally feasible for kinodynamic systems, even in high-dimensional state spaces. We plan to make the source code of our implementation publicly available for download.

For our experiments, we have not fully optimized our implementation, and we believe that running times can be further improved. In particular, extensions suggested in earlier work \cite{karaman11}, such as using an admissible heuristic that can be quickly computed and provides a conservative estimate of the true cost of moving between two states may prune many (relatively costly) attempts to connect pairs of states. Such a heuristic can then also be used in a branch-and-bound technique \cite{karaman11} to prune parts of tree of which one knows it will never contribute to an optimal solution. Further, the constant involved in the rate by which the neighbor radius is allowed to decrease is difficult to estimate for kinodynamic systems, which prompted us to use very conservative radii. Further analysis of the reachable set is needed to establish reasonable estimations. This would potentially also aid in developing a form of efficient neighbor searching for non-Euclidean state spaces (we currently use a brute-force approach), which is still largely an unexplored area. 

Other areas of potential improvement include studying non-uniform sampling to accelerate the convergence to optimal solutions. One could sample more heavily around the current optimal solution, or use stochastic techniques to infer distributions of samples that are likely to contribute to an optimal trajectory. For a quadrotor helicopter for instance, one can imagine that there is a strong correlation between its velocity and orientation, which should be reflected in the sampling. In addition, we note that, as mentioned in the introduction, the ability to connect any pair of states can be used to perform trajectory smoothing by iterative shortcutting as post-processing step. This may improve the quality of solutions further and provide better estimates of the convergence rate of the algorithm.

Lastly, we plan to apply our planner to real-world robots, in particular quadrotors. This would require constructing a stabilizing controller around the computed trajectory, either using traditional techniques such as LQR, or by repeatedly computing reconnections between the current state of the robot and a state on the trajectory.



\bibliographystyle{plainnat}
\begin{footnotesize}

\end{footnotesize}


\end{document}